\documentclass[a4paper,fleqn]{cas-sc}

\usepackage[numbers]{natbib}
\usepackage{multirow, ulem, amsmath, amssymb}

\def\tsc#1{\csdef{#1}{\textsc{\lowercase{#1}}\xspace}}
\tsc{WGM}
\tsc{QE}
\tsc{EP}
\tsc{PMS}
\tsc{BEC}
\tsc{DE}

\begin{document}
\let\WriteBookmarks\relax
\def\floatpagepagefraction{1}
\def\textpagefraction{.001}

\shorttitle{Towards the Holographic Characteristic of LLMs for Efficient Short-text Generation}

\shortauthors{Shun Qian et~al.}

\title [mode = title]{Towards the Holographic Characteristic of LLMs for Efficient Short-text Generation}          
\tnotemark[1]

\tnotetext[1]{The research in this article is supported by the National Key Research and Development Project (2021YFF0901600), 
National Natural Science Foundation of China (62176074), 
Interdisciplinary Development Program of Harbin Institute of Technology(No. SYL-JC-202203)
and the Fundamental Research Funds for the Central Universities (project number: 2022FRFK0600XX).}

\author[1]{Shun Qian}[type=author, style=chinese,
                        orcid=0000-0002-2443-2954]



\ead{shunqian@insun.hit.edu.cn}


\credit{Conceptualization, Methodology, Software, Investigation, Writing – original draft}

\affiliation[1]{organization={Harbin Institute of Technology},
    city={Harbin},
    postcode={150001}, 
    country={Chinese}}

\author[1]{Bingquan Liu}[style=chinese]
\cormark[1]

\ead{bqliu@hit.edu.cn}

\credit{Conceptualization, Methodology, Software, Supervision, Writing – review and editing}

\author[1]{Chengjie Sun}[style=chinese,
   ]


\credit{Software, Supervision, Writing – review and editing}



\author[1]{Zhen Xu}[style=chinese]
\credit{Software, Supervision, Writing – review and editing}

\author[2]{Baoxun Wang}[style=chinese]
\credit{Software, Supervision, Writing – review and editing}

\affiliation[2]{organization={Independent Researcher},
    city={Beijing},
    postcode={100080}, 
    country={Chinese}}

\cortext[cor1]{Corresponding author}

\begin{abstract}
The recent advancements in Large Language Models (LLMs) have attracted interest in exploring their in-context learning abilities and chain-of-thought capabilities. 
However, there are few studies investigating the specific traits related to the powerful generation capacity of LLMs. 
This paper aims to delve into the generation characteristics exhibited by LLMs. 
Through our investigation, we have discovered that language models tend to capture target-side keywords at the beginning of the generation process. 
We name this phenomenon the Holographic Characteristic of language models.
For the purpose of exploring this characteristic and further improving the inference efficiency of language models, we propose a plugin called HOLO, which leverages the Holographic Characteristic to extract target-side keywords from language models within a limited number of generation steps and complements the sentence with a parallel lexically constrained text generation method.
To verify the effectiveness of HOLO,
we conduct massive experiments on language models of varying architectures and scales in the short-text generation scenario. 
The results demonstrate that HOLO achieves comparable performance to the baselines in terms of both automatic and human-like evaluation metrics and highlight the potential of the Holographic Characteristic.
\end{abstract}

\begin{keywords}
Large Language Models \sep Generation capability \sep Holographic Characteristic \sep Short-text Generation \sep Inference Acceleration
\end{keywords}

\maketitle

\section{Introduction}

Recently,
the great progress of Large Language Models (LLM) has brought new End-to-End uniform solutions to almost all the NLP
tasks~\cite{chowdhery2022palm,zhang2022opt,zeng2022glm,touvron2023llama,bubeck2023sparks,anil2023palm}. 
Moreover,
LLMs have exhibited plenty of impressive capabilities, such as In-context Learning~\cite{xieexplanation} and Chain-of-Thought~\cite{wei2022chain,fu2022does}.
Numerous studies~\cite{ wei2022emergent, lu2023emergent, schaeffer2024emergent} are conducted to delve into how LLMs obtain these emergent abilities and how to further enhance and enrich such capabilities.



Even though the generation capability is the core capability of LLMs, the research on it primarily focuses on improving the inference efficiency because of the autoregressive (AR) generative manner~\cite{gu2018non, chen2023accelerating, leviathan2023fast}.
Some works~\cite{xiao2023smoothquant,kim2023squeezellm,kwon2023pageattention} improve the efficiency by reducing the memory consumption of LLMs.
Some other works~\cite{chen2023accelerating, leviathan2023fast,miao2023specinfer} focus on minimizing the number of decoding steps with the speculative sampling strategy.
However,
there are few studies exploring the essence of the generation process of LLMs and further leveraging its characteristics to speed up the inference procedure.

Interestingly, it can be observed that some keywords are assigned higher probabilities in the early stages of generation, although not always the highest probabilities.
This phenomenon, called the Holographic Characteristic of language models, is partially supported by earlier studies that aimed to avoid safe responses by starting generation from the middle keywords~\cite{dinanwizard,zhu2022kpt}.
To demonstrate the Holographic Characteristic more directly, we present a group of results in Table~\ref{table: keywords cover rate}. 
It is evident that over 40\% of the target-side keywords, extracted by a simple statistical method from the beginning decoding step, are present.
Given this observation, we wonder whether it is possible to leverage the Holographic Characteristic to improve the inference efficiency of LLMs.

\begin{table}
\centering
\begin{tabular}{l|lll}
\hline
\textbf{Method} & \textbf{Douban} & \textbf{Weibo} & \textbf{LCCC}\\
\hline
EVA2.0-2.8B & 61.2\% & 56.6\% & 65.0\% \\
ChatGLM-6B & 42.7\% & 46.4\% & 49.6\% \\
Belle-13B & 44.9\% & 48.6\% & 55.1\% \\
\hline
\end{tabular}
\caption{\label{table: keywords cover rate} The proportion of target keywords captured by the vocabulary distribution of the first inference step of different scale LLMs on three datasets.
For each sample, the ground-truth keywords are extracted from generated sentences by LLMs and tokens in the top 1\% probability are selected as candidate keywords.
}
\end{table}

Consequently, we propose a generation plugin called HOLO to leverage the strong generation capabilities of LLMs and generate text in parallel. 
Unlike existing methods for accelerating LLM inference, the HOLO plugin does not require a smaller draft model like speculative sampling~\cite{chen2023accelerating, leviathan2023fast} or training a new model based on the strong conditional independence assumption like non-autoregressive generation (NAR) methods~\cite{huang2022cmlmc, xiao2023survey}. 
It does not make any modifications to the parameters or architecture of the LLMs. 
Specifically, the HOLO plugin follows a two-step process for sentence generation. 
Firstly, it utilizes the Holographic Characteristic to extract target-side keywords by analyzing the probability distributions of the first two generation steps of LLMs. 
Secondly, it employs the lexically constrained text generation method POINTER~\cite{zhang2020pointer} to complete sentences based on the extracted results.

Experimental results on dialogue generation tasks using multiple language models as the base models show that the HOLO plugin can generate coherent and unique sentences.
The quality of sentences generated by HOLO is comparable to those produced by the base models.
This performance provides strong evidence of the effectiveness of the Holographic Characteristic exhibited by the language models.
Additionally, the utilization of the HOLO plugin greatly improves the inference efficiency of the base language models.

\section{Related Work}

\subsection{Non-Autoregressive Generation}
To tackle the inference efficiency problem of AR models, the non-autoregressive generation approach is proposed to accelerate the decoding procedure of text generation.
On the translation task, NAT~\cite{gu2018non} is first proposed to generate all the target tokens parallelly based on the conditional independence assumption as shown in Eq.~\ref{nar}.
\begin{equation}
\label{nar}
    P_{NAR}(Y|X) = \prod_{i=1}^{T}P(y_i|X)
\end{equation}
where each token $y_i$ only relies on the context sequence $X$ and has no dependency on other tokens in $Y$.
This work has notably increased the inference speed, 
and thus is followed by a series of studies~\cite{ghazvininejad2019mask-predict,dingunderstanding,wang2019non}.
Some works introduced the latent variable to model the dependency between target-side tokens implicitly~\cite{ran2021reordernat,bao2021cnat, bao2022glat}.
Other works explored to refine the generated token sequence iteratively instead of generating the target sentence in one pass~\cite{ghazvininejad2019mask-predict,saharia2020ctc, huang2022cmlmc}.
The challenge of applying the NAR technique on LLMs lies in that the generation models have to be optimized 
under the independence assumption which is resource-consuming for LLMs.

\subsection{LLM Inference Acceleration}
There are two types of research works proposed to alleviate the inefficiency of LLM inference.
One line of research aims to reduce memory consumption, thereby minimizing memory transfer overhead and enabling large batch size~\cite{cai2024medusa}.
Grouped-query attention~\cite{ainslie2023gqa} and PagedAttention~\cite{kwon2023pageattention}
substantially cut the memory consumption of the KV cache.
Quantization techniques are also used to shrink the memory consumption of LLMs~\cite{frantar2022gptq,xiao2023smoothquant,kim2023squeezellm}.
Another line of research pays attention to minimizing the number of decoding steps with the speculative sampling strategy.
Speculative sampling~\cite{chen2023accelerating, leviathan2023fast} employs a smaller draft model to conjecture several subsequent words, the LLMs parallel evaluate these words and accept them as appropriate.
Tree-structured attention~\cite{miao2023specinfer,spector2023accelerating} is further proposed to generate multiple candidates in parallel.
Different from these works, our HOLO plugin only relies on the distribution of the first two decoding steps of LLMs and makes no change to the parameters of LLMs.


\section{Methods}
\label{sec:method}

Given a context, the Holographic Characteristic of LLMs tends to capture target-side keywords at the very beginning of the generation process.
Based on this characteristic, the proposed HOLO plugin extracts the target-side keywords from the distribution of the first two inference steps of LLMs.
Then, HOLO employs a modified lexically constrained text generation approach to produce a natural sentence based on obtained keywords.


\subsection{Target-side Keywords Extraction}

Given an input sequence $X$, the probability of target-side word $w$ can be formulated as:
\begin{equation}
\label{eq_w_on_x}
    P(w|X) =  \frac{1}{N}\sum_{i=1}^{N} P(w, pos=i|X) =  \frac{1}{N}\sum_{i=1}^{N} P(y_i=w|X)
\end{equation}
where $N$ is the length of target sequence $Y$, $y_i=w$ denotes that $w$ appears in the $i$-th position of $Y$.
Thus, if the probability of $w$ appearing at each position in sequence $Y$ is known,
it will be trivial to get $P_{\mathcal{F}}(w|X)$ for model $\mathcal{F}$.
Unfortunately, language models usually depend on the chain rule of conditional probability $P(y_i|y_{<i}, X)$, it is difficult to get $P(y_i=w|X), i>1$.

To address the above issue, we propose a novel method for estimating $P_{\mathcal{F}}(w|X)$ using the distribution of the first two generation steps based on the Holographic Characteristic.
Specifically, we introduce a simple first-order Markov assumption: given the context $X$, the probability of word $y_i$ can be approximated only relying on the previous word:
\begin{equation}
    P(y_i|y_{<i},X) \approx P(y_i|y_{i-1},X)
\end{equation}
Based on this assumption, $P(y_i|y_{i-1},X)$ can be approximated as below:
\begin{equation}
\label{equ:markov_assumption}
    P(y_i|y_{i-1},X) = P(y_2|y_1=y_{i-1},X)
\end{equation}
Thus, we can estimate $P_{\mathcal{F}}(y_i|X)$ as follow:
\begin{equation}
\label{chain_rule}
    \begin{aligned}
               P_{\mathcal{F}}(y_i|X) 
        = \sum_{y_{i-1} \in V}P_{\mathcal{F}}(y_i|y_{i-1},X)\cdot P_{\mathcal{F}}(y_{i-1}|X)  \approx \sum_{y_{i-1} \in V}P_{\mathcal{F}}(y_2|y_1 = y_{i-1},X)\cdot P_{\mathcal{F}}(y_{i-1}|X)
    \end{aligned}
\end{equation}

where $V$ denotes the vocabulary of the model $\mathcal{F}$.

As a result, the inference procedure of target sequence $Y$ is a Markov process.
The initial state distribution $\pi_0 = (\pi_0(1), \pi_0(2), ..., \pi_0(|V|))^T$ of the Markov process is: 

\begin{equation}
    \pi_0(j) \approx P_{\mathcal{F}}(y_1=v_j|X)
\end{equation}

where $v_j$ is the $j$-th word in vocabulary $V$ and $|V|$ is the vocabulary size.
The transition matrix $\mathcal{M} = (m_{jk}) \in \mathbb{R}^{|V| \times |V|}$ of the Markov process is:
    \begin{equation}
        m_{jk} \approx P_{\mathcal{F}}(y_2=v_j|y_1=v_k,X)
    \end{equation}
At last, the i-th step state distribution $P(y_i|X), i\in [1,N]$ can be formulated as:
    \begin{equation}
        P_{\mathcal{F}}(y_i|X) \approx M^{i-1} \pi_0
    \end{equation}


However, since the vocabulary size of language models usually ranges from 50,000 to 100,000, it is a non-trivial task to estimate $\mathcal{M}$ and $\pi_0$.
Fortunately, given a context $X$, the semantic of $Y$ is constrained into a small space, in which finite keywords would appear in $Y$.
In other words, for a given sequence $X$, the vocabulary of keywords (denoted by $V^y$) that may appear in the generated sequence $Y_{\mathcal{F}}$ must be much smaller than $|V|$. 
What's more, the Holographic Characteristic of LLMs denotes that target-side keywords will be assigned higher probabilities than other words.
Thus, the probability of the small set $V^y$ may occupy the most likelihood of $\pi_0$.
Inspired by the nucleus sampling, we heuristically take the $V_1^{(p)}$ as the approximate of $V^y$, where $V_1^{(p)}$ is the smallest set of tokens satisfying to

\begin{equation}
\label{eq: vp}
    \sum_{w\in V_1^{(p)}}P_{\mathcal{F}}(y_1=w|X) \ge p
\end{equation}

Then, the distribution of $P_{\mathcal{F}}(y_2|X)$ can be divided into two parts:
\begin{equation}
\label{eq: y2}
    \begin{aligned}
         P_{\mathcal{F}}(y_2|X) 
         = & \sum_{y_1 \in V_1^{(p)}}P_{\mathcal{F}}(y_2|y_1,X)\cdot P_{\mathcal{F}}(y_{1}|X) + \sum_{y_1 \in V - V_1^{(p)}}P_{\mathcal{F}}(y_2|y_1,X)\cdot P_{\mathcal{F}}(y_{1}|X) \\
        \leq & \sum_{y_1 \in V_1^{(p)}}P_{\mathcal{F}}(y_2|y_1,X)\cdot P_{\mathcal{F}}(y_{1}|X)  + (1-p)\sum_{y_1 \in V - V_1^{(p)}}P_{\mathcal{F}}(y_2|y_1,X) \\
    \end{aligned}
\end{equation}

In practice, since the size of $V$ is always bigger than 50,000 and the size of $V^y$ is commonly less than 100,
the value of $\sum_{y_1 \in V - V_1^{(p)}}P_{\mathcal{F}}(y_2|y_1,X)$ is close to the constant $\frac{|V| - |V_1^{(p)}|}{|V|}$.
If $p$ is close to 1, 
the second item of Eq.~\ref{eq: y2} would be less than a small constant $\mathcal{C}$.
In this case, estimating $P_{\mathcal{F}}(y_2|X)$ using $\sum_{y_1 \in V_1^{(p)}}P_{\mathcal{F}}(y_2|y_1,X)\cdot P_{\mathcal{F}}(y_{1}|X)$ contains a bias less than $\mathcal{C}$.

Following the above approximating procedure, we can further estimate $P_{\mathcal{F}}(y_i|X), i>2$ by Eq.~\ref{chain_rule}.
Although with the increase of the position $i$, the estimation bias will also inevitably increase, the bias is still tiny since $V^{(p)}$ dominates the most probability of the generation procedure.
Therefore, with the Holographic Characteristic of LLMs, $P_{\mathcal{F}}(y_i|X), i>2$ can be approximated using the states fo the first two steps of the generation procedure.
In practice, we estimate $P_{\mathcal{F}}(w|X)$ as follows:
\begin{equation}
    \begin{aligned}
        &P_{\mathcal{F}}(w|X) &\approx \frac{1}{T}\sum_{t=1}^T p_{\mathcal{F}}(y_i|X) \\
    \end{aligned}
\end{equation}
where $T$ is the step of the generation procedure.
Afterward, we can extract the target-side keywords based on $P_{\mathcal{F}}(w|X)$.
Specifically, the intersection of the set of top-k probable words and the initial set $V_1^{(p)}$ is selected as the final $V_{\mathcal{F}}^y$.

\subsection{Modified Lexically Constrained Text Generation}
\label{sec:modified pointer}

\begin{table*}[!htbp]
    \centering
        
    \begin{tabular}{l|l}
         \hline
         \textbf{Description} & \textbf{Examples} \\
         \hline
         input context $X$ & I'm on the beach in Los Angeles, the water is so cold! \\
         \hline 
         \multirow{2}{*}{keywords $V^y$} & Los Angeles, weather, fish, sea, beach, warm, swim, summer \\ 
         & winter, city, cool, American, ... \\
         \hline 
         \multirow{1}{*}{keyword chains} & cool-summer-weather-warm-swim-sea-Los Angeles, \\ 
         & American-swim-beach-sea-fish, ... \\
         \hline
         POINTER Input ($Y^0$)  & cool summer weather warm swim sea Los Angeles \\
         \hline
         \multirow{2}{*}{POINTER $Y^1$} & \textcolor{blue}{It's(0.65)} cool \textcolor{blue}{.(0.48)} summer \sout{\textcolor{blue}{sunshine(0.16)}} \textcolor{blue}{when(0.32)} weather  \\
         & \sout{\textcolor{blue}{very(0.12)}} warm \textcolor{blue}{,(0.35)} swim \textcolor{blue}{in(0.44)} sea \textcolor{blue}{of(0.73)} Los Angeles \\
         \hline
         \multirow{3}{*}{POINTER $Y^2$} & \textcolor{blue}{It's} cool \textcolor{blue}{.} \textcolor{magenta}{Every(0.37)} summer \textcolor{magenta}{,(0.25)} \textcolor{blue}{when} \textcolor{magenta}{the(0.53)} weather \\ 
         & \textcolor{magenta}{become(0.49)}  warm \textcolor{blue}{,} \textcolor{magenta}{people(0.22)} swim \textcolor{blue}{in} \sout{\textcolor{magenta}{a(0.19)}} sea \textcolor{blue}{of} \\
         & Los Angeles \textcolor{magenta}{.(0.57)} \\
         \hline
         \multirow{3}{*}{POINTER $Y^3$} & \textcolor{blue}{It's} cool \textcolor{blue}{.} \textcolor{magenta}{Every} summer \textcolor{magenta}{,} \textcolor{blue}{when} \sout{\textcolor{olive}{,(0.07)}} \textcolor{magenta}{the} weather \textcolor{magenta}{become}  \\ 
         & \sout{\textcolor{olive}{much(0.12)}} warm \textcolor{olive}{er(0.36)} \textcolor{blue}{,} \sout{\textcolor{olive}{see(0.16)}} \textcolor{magenta}{people} \textcolor{olive}{will(0.63)} swim \textcolor{blue}{in} \\
         & \textcolor{olive}{the(0.36)} sea \textcolor{blue}{of} Los Angeles \textcolor{magenta}{.} \\
         \hline
         \multirow{2}{*}{POINTER Output } & \textcolor{blue}{It's} cool \textcolor{blue}{.} \textcolor{magenta}{Every} summer \textcolor{magenta}{,} \textcolor{blue}{when} \textcolor{magenta}{the} weather \textcolor{magenta}{become} warm \textcolor{olive}{er} \textcolor{blue}{,}   \\ 
         & \textcolor{magenta}{people}  \textcolor{olive}{will} swim \textcolor{blue}{in} \textcolor{olive}{the} sea \textcolor{blue}{of} Los Angeles \textcolor{magenta}{.} \\
         \hline
         
    \end{tabular}
\caption{The generation process of POINTER with mask-predict strategy. Words followed by bracketed prediction probability indicate newly generated words at the current stage.
    Words with \sout{\textcolor{red}{strikeout}} denote the low-confidence ones and will be eliminated for better quality. }
    \label{table:pointer}
\end{table*}

\subsubsection{Building Keyword Chain}
\label{sec:word-chain}




Once the target-side keywords are obtained, it is crucial to organize these unordered keywords into ordered keyword chains. 
This step is essential in addressing the issue of lacking target-side dependency, which is the primary reason for the unsatisfactory performance of existing NAR methods.
Besides,
the word chain is helpful for generating high-quality sentences and improving generation efficiency.


Consequently, a beam-search-style strategy is proposed to build keyword chains based on the model-specific conditional probability $P_{\mathcal{F}}(y_2|y_1,X), y_1\in V_1^{(p)}$.
%
For a chosen keyword $w \in V_{\mathcal{F}}^y$, we select top-k keywords as the next step candidates according to the conditional probability $P_\mathcal{F}(y_2|y_1=w, X)$.
Repeat the above step until the chain length is up to a pre-defined maximum length $L$ or the probability of the keyword chain is lower than a threshold.
The probability of the keyword chain $C=\{c_1, c_2, ..., c_m\}$ is defined as:
\begin{equation}
      P_\mathcal{F}(C|X)  =  P_\mathcal{F}(y_1=c_1|X) \cdot\prod_{i=2}^{m}P_\mathbf{F}(w=c_i|X)P_\mathcal{F}(y_2=c_i|y_1=c_{i-1},X)
\end{equation}
At last, we pick $Z$ chains with the highest probability $P_\mathcal{F}(C|X)$ for further generation.



\subsubsection{Lexically Constrained Text Generation Using Modified POINTER}

After obtaining the keyword chain $C$, the next step is to generate a sentence based on it.
To ensure that the generated sentence contains most keywords of $C$ and maintains the relative position, we adopt POINTER~\cite{zhang2020pointer}, a simple yet novel iterative insertion-based parallel method, as our lexically constrained text generation model. 

Given lexical constraints, POINTER first generates high-level words that bridge
the keyword constraints, then these words are used as pivoting points at which to insert details of finer granularity. This process iterates until a sentence is finally completed by adding the least informative words (typically pronouns and prepositions).
Specifically,
the keyword chain $C$ is taken as the initialization of the generation procedure (denoted by $Y^0$).
The generation procedure of POINTER can be formulated as a sequence of $K$ stages:
$S=\{Y^0, Y^1, ..., Y^{K-1}, Y^{K}\}$.
For each $k \in \{1, ..., K\}$, $Y^{K-1}$ is a sub-sequence of $Y^K$.
In other words, the following stage can be perceived as a finer-resolution text sequence compared to the preceding stage.
$Y^K$ is the final generation, under the condition that the iterative procedure is converged (i.e., $Y^{K-1}=Y^K$).
Formally, the distribution can be factorized as follows:
\begin{equation}
    p(Y)=p(Y^0)\prod_{k=1}^K p(Y^k|Y^{k-1})
\end{equation}
where $p(Y^k|Y^{k-1})=\prod_{y\in Y^k-Y^{k-1}} p(y|Y^{k-1})$.

Besides, we incorporate a simple but effective strategy \textit{mask-predict}~\cite{ghazvininejad2019mask-predict} into the insertion-based iterative procedure to alleviate the low-quality issue of generated texts,
which is caused by the forcible insertion policy of POINTER.
The mask-predict policy eliminates words with low generation probability in the current generated sentence, and re-predicts them later with more information.
With this strategy, POINTER is able to reconsider the token choices, so as to generate high-quality pieces in future steps.
Table~\ref{table:pointer} shows an example of the iterative text generation process.

Overall, 
since multiple keyword chains are selected for each sample,
and POINTER with mask-predict strategy generates one candidate sentence for each keyword chain.
Thus, a ranker is required to rate these generated sentences to select the best one for the given context.
The implementation of the ranker is detailed in Section~\ref{subsec:imp_detail}.


\section{Experiments}
To verify the Holographic Characteristic of language models,
the performance and generation efficiency of the HOLO plugin is evaluated on the short-text generation task.
The remaining part of this section describes our experimental setup (e.g., baselines, datasets, implementation details), results and discussion in detail.




\subsection{Experimental Setup}

\subsubsection{Evaluation Datasets}
A classic short-text task, open-domain dialogue generation, is adopted to validate the generation capability of our HOLO plugin.
Since the HOLO is a language model-based generation plugin and no pre-training is needed, we evaluate HOLO on the testing sets of three Chinese dialogue datasets: Douban~\cite{wu2017douban}, Weibo~\cite{shang2015weibo} and LCCC~\cite{wang2020lccc}, which consist of 1,000, 10,000, and 10,000 examples respectively.


\subsubsection{Baselines}
Three language models with different architectures and parameter scales are chosen as the base models of the HOLO plugin.
The details of these models are as follows:
\begin{itemize}
    \item EVA2.0-2.8B~\cite{gu2023eva2}: it is the largest Chinese open-source pre-trained dialogue model trained on 400 million high-quality dialogue samples.
    \item ChatGLM-6B~\cite{zeng2023glm-130b}: it is an open bilingual language model and pre-trained with 1T tokens of Chinese and English corpus, supplemented by supervised fine-tuning, feedback bootstrap, and reinforcement learning with human feedback.
    \item Belle-13B~\cite{BELLE}: it is based on LLaMA~\cite{touvron2023llama} and secondary pre-trained with 3.4 billion Chinese words for improving the performance of LLaMA in the Chinese domain.
\end{itemize}


\subsubsection{Implementation Details}
\label{subsec:imp_detail}
The value of $p$ introduced in Eq.~\ref{eq: vp} is set to 0.9 and the picked keyword chain number $Z$ is set to 5. 
The pre-defined keyword chain maximum length $L$ is set to 7/12/10 for EVA2.0-2.8B/ChatGLM-6B/Belle-13B based on the average keywords number of generated sentences on the Douban dataset.

As described in Section~\ref{sec:modified pointer}, a ranker is required to select high-quality generated sentences.
Following the work of~\cite{gu2023eva2},
we fine-tune a BERT~\cite{devlin2018bert} model to assess the semantic relevance between the given content and the response.
The P@1/10 of the ranker is 0.86 evaluated on 10,000 test samples.
Besides, due to only the English POINTER model being released~\footnote{https://github.com/dreasysnail/POINTER},
we pre-train a Chinese POINTER model following the same pre-training setup. 
The pre-training details of the Chinese POINTER model are provided in Appendix~\ref{appendix:ranker}.

\subsection{Experiment Results}

To assess the quality of sentences generated by the HOLO plugin and its base models, we employ both automatic and human evaluations, similar to the majority of dialogue generation studies~\cite{zhang2019dialogpt, chen2023diag1, bao2021plato}. 

\begin{table*}[!htbp]
\centering

\begin{tabular}{llccccccc}
\hline
\textbf{Dataset} & \textbf{Method} & \textbf{F1} & \textbf{Rouge-L} & \textbf{BLEU-2} & \textbf{BLEU-4} & \textbf{Distinct-2}  & \textbf{PPL} & \textbf{Rel.}  \\
\hline
\multirow{6}{*}{\textbf{Douban}} & 
EVA2.0-2.8B & 9.59 & 7.62 & 0.95 & 0.17 & 34.22 & 135.12 & 0.61  \\
& HOLO$_{\rm EVA2.0-2.8B}$ & \textbf{9.87} & \textbf{7.89} & \textbf{2.21} & \textbf{0.48} & \textbf{56.22} & \textbf{120.58} & \textbf{0.73} \\
\cline{2-9}
& ChatGLM-6B & 9.69 & 7.67 & 3.63 & \textbf{0.88} & 34.79  & \textbf{31.16} & 0.79\\
& HOLO$_{\rm ChatGLM-6B}$ & \textbf{10.71} & \textbf{8.05} & \textbf{3.65} & 0.47 & \textbf{42.83} & 44.33 & \textbf{0.81}  \\
\cline{2-9}
& Belle-13B & 8.84 & 6.77 & 3.32 & \textbf{0.51} & \textbf{39.26} & \textbf{37.69} & 0.53  \\
& HOLO$_{\rm Belle-13B}$ & \textbf{11.02} & \textbf{7.97} & \textbf{3.50} & 0.45 & 31.71 & 42.66 & \textbf{0.67}  \\ 
\hline

\multirow{6}{*}{\textbf{Weibo}} & 
EVA2.0-2.8B & \textbf{12.94} & \textbf{10.72} & 4.60 & 1.08 & 21.50 & 179.54 & 0.63 \\
& HOLO$_{\rm EVA2.0-2.8B}$ & 10.89 & 9.34 & \textbf{4.74} & \textbf{1.11} & \textbf{29.86} & \textbf{104.29} & \textbf{0.73} \\
\cline{2-9}
& ChatGLM-6B & 9.78 & 8.65 & 3.23 & \textbf{1.23} & 19.08 & \textbf{32.21} & 0.74 \\
& HOLO$_{\rm ChatGLM-6B}$ & \textbf{11.98} & \textbf{9.98} & \textbf{3.80} & 0.71 & \textbf{21.16} & 42.57 & \textbf{0.81} \\
\cline{2-9}
& Belle-13B & 11.28 & 9.61 & \textbf{2.93} & \textbf{0.72} & \textbf{18.07} & 42.43 & 0.61 \\
& HOLO$_{\rm Belle-13B}$ & \textbf{12.31} & \textbf{10.03} & \textbf{3.64} & 0.60 & 15.35 & \textbf{39.56} & \textbf{0.72} \\
\hline

\multirow{6}{*}{\textbf{LCCC}} & 
EVA2.0-2.8B & \textbf{11.75} & \textbf{10.42} & \textbf{4.23} & \textbf{1.03} & 19.11 & 194.48 & 0.58 \\
& HOLO$_{\rm EVA2.0-2.8B}$ & 10.12 & 9.18 & \textbf{3.62} & 0.62 & \textbf{24.57} & \textbf{108.27} & \textbf{0.67} \\
\cline{2-9}
& ChatGLM-6B & 7.26 & 6.69 & 1.91 & 0.39 & 16.49 & \textbf{34.89} & 0.57 \\
& HOLO$_{\rm ChatGLM-6B}$ & \textbf{10.10} & \textbf{8.94} & \textbf{2.93} & \textbf{0.44} & \textbf{19.35} & 44.48 & \textbf{0.68} \\
\cline{2-9}
& Belle-13B & 9.73 & 8.68 & 2.54 & \textbf{0.41} & \textbf{18.99} & \textbf{33.13} & 0.47 \\
& HOLO$_{\rm Belle-13B}$ & \textbf{10.05} & \textbf{8.69} & \textbf{2.60} & 0.36 & 13.43 & 41.13 & \textbf{0.59} \\
\hline
\end{tabular}
\caption{Automatic evaluation results of three different parameter scale LLMs on three datasets. 
``R-L/B-2/B-4/D-2/Rel.'' stands for ``ROUGE-L/BLEU-2/BLEU-4/Dist-2/Relevance''.
The better results of each group are highlighted in bold. 
The item ``HOLO$_{\rm XXX}$'' refers that taking ``XXX'' as the base model of our HOLO plugin.}
\label{table:metrics}
\end{table*}

\subsubsection{Results of Automatic Evaluation}


The most popular automatic metrics including uni-gram F1, ROUGE-L~\cite{lin2004rouge_l}, BLEU~\cite{papineni2002bleu}, Distinct 2-grams~\cite{li2015distinct} and GPT-2 Perplexity (PPL) are adopted to estimate the quality of the generated text. 
The automatic evaluation results of the HOLO plugin and its corresponding three base models on the three datasets are presented in Table~\ref{table:metrics}.


As shown in Table~\ref{table:metrics}, the HOLO plugin outperforms the corresponding base models on 5 out of 7 metrics, including F1, ROUGE-L, BLEU-2, Dist-2, and Rel.
This result validates the effectiveness of the HOLO plugin in short-text generation task. 
Furthermore, it is worth noting that the HOLO plugin does not perform as well as the base models on BLEU-4 and PPL.
This observation can be attributed to the inherent limitations of the parallel generative model POINTER and highlights the fact that the lexically constrained text generation approach is a bottleneck for the HOLO plugin.


In addition to the standard generation metrics, we also utilize the semantic relevance score provided by the fine-tuned BERT model mentioned above as an alternative perspective to evaluate the relevance between the given context and the generated text.
Table~\ref{table:metrics} shows that the HOLO plugin significantly outperforms its base model on the relevance metric Rel.
This shows the HOLO plugin's capability to generate reasonable responses for short-text generation task.

On the whole, the performance of the HOLO plugin on these automatic evaluation metrics  is significantly consistent with that of the base models.
In other words, the better the base model performs, the better the HOLO plugin performs.
For example, 
on the Douban dataset, ChatGLM-6B performs better than EVA2.0-2.8B on all the metrics,
the HOLO plugin taking ChatGLM-6B as the base model performs better than that taking EVA2.0-2.8B as the base model on most automatic metrics.
That shows that the performance of HOLO is constrained by the capability of its base model.

\subsubsection{Results of Human-like Evaluation}

Following~\cite{bao2021plato}, we perform a human evaluation from three perspectives, namely informativeness, coherence, and humanness (details are described in Appendix~\ref{appendix: geval}).
Instead of manually assessing the sentence quality with human bias, 
we employ the LLM-based G-Eval~\cite{liu2023geval} method to assign scores on the three metrics using a scale of 1-5. 
Specifically, G-Eval utilizes GPT-4 with Chain-of-Thought (CoT) prompts to evaluate the quality of generated text, achieving a high level of correspondence with human evaluations. The CoT prompts and more details are provided in Appendix~\ref{appendix: geval}.
Additionally, we apply G-Eval on 200 randomly sampled dialogues from each dataset.
The human evaluation results on the three datasets are shown in Table~\ref{table:human}.

It can be seen that the models incorporating the HOLO plugin achieve comparable performance to their corresponding base models.
Nevertheless, there is a slight decrease in three human-like evaluation metrics for ChatGLM-6B and Belle-13B. 
Specifically, the average score reductions for ChatGLM-6B in Informativeness, Coherence, and Humanness are 0.15, 0.16, and 0.32, respectively.
And the performance reductions of Belle-13B in the three human evaluation metrics are 0.16, 0.23, and 0.27, respectively.
The slight decrease in Informativeness and Coherence highlights the HOLO's ability to generate informative and dialogue-coherent text while the decrease in Humanness is more pronounced, indicating a drawback in the lexically constrained generation approach when it comes to generating human-like and natural language sentences.
Noting that the performance of HOLO based on EVA2.0 is significantly better than the base model and reduces the gap between EVA2.0 and ChatGLM-6B or Belle-13B. 
This observation shows that HOLO is capable of mitigating the architecture and pre-training data-size disadvantages of EVA2.0.

\begin{table*}[!htbp]
    \centering
    \begin{tabular}{l|ccc|ccc|ccc}
    \toprule
    \multirow{2}{*}{\textbf{Method}  } &
    \multicolumn{3}{c}{\textbf{Douban}} & 
    \multicolumn{3}{|c|}{\textbf{Weibo}} &
    \multicolumn{3}{c}{\textbf{LCCC}} \\
    \cline{2-10}
    & Inf. & Coh.  & Hum. & Inf. & Coh.  & Hum. & Inf. & Coh.  & Hum. \\
    \midrule
    EVA2.0-2.8B & 2.50 & 2.38 & 2.80 & 2.75 & 2.84 & 2.68 & 2.83 & \textbf{3.22} & 3.08 \\
    HOLO$_{\rm EVA2.0-2.8B}$ & \textbf{2.87} & \textbf{2.52} & \textbf{3.12} & \textbf{3.06} & \textbf{2.91} & \textbf{3.37} & \textbf{3.23} & 3.20 & \textbf{3.46} \\
    \hline
    ChatGLM-6B & \textbf{3.44} & \textbf{3.71} & \textbf{3.98} & \textbf{3.39} & 
    \textbf{3.47} & \textbf{3.93} & \textbf{3.26} & \textbf{3.54} & \textbf{3.80} \\
    HOLO$_{\rm ChatGLM-6B}$ & 3.32 & 3.49 & 3.53 & 3.20 & 3.35 & 3.71 & 3.12 & 3.41 & 3.52\\
    \hline
    Belle-13B & \textbf{3.26} & \textbf{3.46} & \textbf{3.69} & \textbf{3.21} & \textbf{3.61} & \textbf{3.57} & \textbf{3.42} & \textbf{3.59} & \textbf{3.72}\\
    HOLO$_{\rm Belle-13B}$ & 3.15 & 3.22 & 3.27 & 3.07 & 3.39 & 3.41 & 3.19 & 3.36 & 3.48\\
    \bottomrule
    \end{tabular}
    \caption{Human evaluation results of the different parameter scale LLMs on three datasets. The results are evaluated by GPT-3.5 with corresponding CoT prompts following the G-Eval method~\cite{liu2023geval}. The evaluation score ranges from 1 to 5. 
    ``Inf./Coh./Hum.'' stands for the Human evaluation metric ``Informativeness/Coherence/Humanness'' respectively.
    The better results of each group are highlighted in bold.
    }
    \label{table:human}  
\end{table*}

Although there is room for improving the lexically constrained text generation approach of the HOLO plugin, 
the performance on shot-text generation still demonstrates the feasibility of generating reasonable responses using only the first-two generation steps of LLMs in short-text generation tasks.
Furthermore, this performance on the dialogue generation task further validates the Holographic Character of LLMs.

\begin{table}[!htbp]
\centering
\begin{tabular}{l|ll}
\hline
\textbf{Method} & \textbf{Time (s)} & \textbf{Memory (MB)} \\
\hline
EVA2.0-2.8B & 0.21 & 2350 \\
HOLO$_{\rm EVA2.0-2.8B}$ & 0.39(+85.7\%) & 1746(-25.7\%) \\ 
\hline
ChatGLM-6B & 6.39 & 13306 \\
HOLO$_{\rm ChatGLM-6B}$ & 0.47(-92.6\%) & 5096(-61.7\%) \\
\hline
Belle-13B & 1.30 & 27170 \\
HOLO$_{\rm Belle-13B}$ & 0.56(-56.9\%) & 12101(-55.5\%) \\
\hline
\end{tabular}
\caption{Average time cost for each sample and GPU memory utilization of three different parameter scale LLMs on the Douban test set.}
\label{table:compute-efficiency}
\end{table}

\subsubsection{The Inference Efficiency Analysis}

During inference,
LLMs usually take the beam search decoding strategy whose time complexity is $\mathcal{O}(NB|V|)$, where $B$ refers to the beam width.
While the HOLO plugin only relies on the first two generation steps, the time complexity of the HOLO plugin is reduced to $\mathcal{O}(|V^y|+log \frac{N}{|C|})$,
where $|C|$ the length of word chain and $|V^y|<<|V|$.

It should be noted that even though the HOLO plugin contains many one-step inferences for capturing target-side keywords, 
these one-step inferences can be calculated in parallel.
Additionally, the rest generation procedure of HOLO is base model-free and has a very low time and memory consumption.
As a result, compared to the base models, the overall time and memory cost after applying the HOLO plugin will be significantly reduced.

To demonstrate the efficiency improvement brought by HOLO,
we measure the time cost and GPU memory usage of the base model and the HOLO plugin during the inference procedure.
The results reported in Table~\ref{table:compute-efficiency} are evaluated on the Douban testing set with a single Nvidia V100 GPU.
It can be seen that the HOLO plugin can maintain the inference time at a low level.
From a time cost perspective, the HOLO plugin reduces time costs by 56.9\% for Belle-13B and 92.6\% for ChatGLM-6B. 
From a memory cost perspective, the HOLO plugin decreases GPU memory usage by 61.7\%/55.5\% for ChatGLM-6B/Belle-13B respectively.

It is worth noting that after applying the HOLO plugin in practice, the time cost of EVA2.0-2.8B increases by 85.7\% even though the GPU memory cost is reduced by 25.7\%.
This phenomenon indicates the uniqueness of the HOLO plugin.
The HOLO plugin relies on the base model to capture the target-side keywords and the rest of the generation procedure is base model-free.
Thus, when the inference speed of the base model is fast, the advantage of the HOLO plugin can not be demonstrated. 
In other words, the slower the base model, the more significant the efficiency improvement  HOLO brings.
Considering the ongoing trend of increasing model scale to get better performance, the meaning of the HOLO plugin will become more important.

\subsection{Is Mask-Predict Strategy Useful?}

\begin{table*}[!htbp]
\centering

\begin{tabular}{lc|ccccc|ccc}
\hline
\textbf{BaseModel} & \textbf{M-P} & \textbf{F1} & \textbf{R-L} & \textbf{B-4}  & \textbf{D-2} & \textbf{PPL}  & \textbf{Inf.} & \textbf{Coh.} & \textbf{Hum.} \\
\hline
\multirow{2}{*}{EVA2.0-2.8B}  &
  $\checkmark$ & 9.87 & 7.89 & \textbf{0.48} & \textbf{56.22} & \textbf{120.58} & \textbf{3.23} & \textbf{3.20} & \textbf{3.46} \\
 & $\times$ & \textbf{10.86} & \textbf{8.72} & 0.42 & 49.43 & 191.80 & 2.68 & 2.41 & 2.76 \\
\hline
\multirow{2}{*}{ChatGLM-6B} & $\checkmark$ & 10.71 & 8.05 & 0.47 & \textbf{42.83} & \textbf{44.33} & \textbf{3.12} & \textbf{3.41} & \textbf{3.52} \\
& $\times$ & \textbf{10.79} & \textbf{8.22} & \textbf{0.48} & 40.82 & 62.81 & 2.81 & 2.87 & 3.01 \\
\hline
\multirow{2}{*}{Belle-13B} & $\checkmark$ &11.02 & 7.97 & \textbf{0.45} & \textbf{31.71} & \textbf{42.66} & \textbf{3.19} & \textbf{3.36} & \textbf{3.48} \\ 
& $\times$ & \textbf{11.32} & \textbf{8.46} & 0.26 & 30.38 & 60.38 & 2.72 & 2.53 & 2.87 \\
\hline
\end{tabular}
\caption{ Automatic \& Human evaluation results on Douban test set. 
``M-P'' stands for ``Mask-Predict''. 
$\checkmark$ indicates \textit{mask-predict} is used and $\times$ indicates not. 
The better results of each group are highlighted in bold. }
\label{table:mask-predict}
\end{table*}

To address the forcible insertion issue of POINTER, 
we integrate the widely-used iterative-based NAR strategy, "mask-predict", into the inference process.
An ablation experiment on the Douban dataset is conducted to verify its effectiveness.
As shown in Table~\ref{table:mask-predict},
after applying the mask-predict strategy, 
even though there are slight performance decreases in F1 and ROUGE-L,
the performance on all other metrics is markedly improved, especially the Coherence and Humanness.
This proves that introducing the mask-predict strategy to the iterative-based POINTER model can indeed improve the quality of generated sentences.

\section{Conclusion}

To explore the generation ability of LLMs,
we propose an efficient short-text generation plugin, HOLO, 
which only relies on the first two generation steps.
The experimental results not only demonstrate the effectiveness of our HOLO plugin but also prove that the beginning steps of LLM generation contain rich target-side semantic information, named as Holographic Characteristic of language models.
Nevertheless, there is a generation discrepancy between LLMs and the HOLO plugin.
We will explore ways to narrow this gap in the future
and we hope this work can promote the understanding of the generation of LLMs.

\appendix

\section{Pre-training Details of the Chinese POINTER Model}
\label{appendix:ranker}
The BaiduBaike dataset is used as the pre-training data.
It contains 5 GB of raw text and is pre-processed into a set of natural sentences, with a maximum sequence length of 128 tokens.
Since the POINTER aims to generate high-level important tokens first and utilizes these tokens as pivoting points for later generations,
it is critical to assess the importance score of each token so that a sentence $X$ can be broken into a consecutive series of pairs $(X^0, X^1), \cdots (X^{K-1}, X^{K})$, where $K$ is the number of such pairs.

The importance score of a token $x_t$ is defined as $\alpha_t = \alpha_t^{TF-IDF} + \alpha_t^{POS} + \alpha_t^{YAKE}$, 
where $\alpha_t^{TF-IDF}$ is the TF-IDF score, 
$\alpha_{POS}$ is the assigned score based on the POS tagging (noun or verb token will be assigned a higher score than other tokens),
$\alpha_t^{YAKE}$ represents whether $x_t$ is the keyword based on the extraction results of YAKE~\cite{Cam_2018_yake} (if $x_t$ is the keyword, $\alpha_t^{YAKE}$ is 1; Otherwise, $\alpha_t^{YAKE}$ is 0).

However, according to our statistics, the variance of TF-IDF scores of Chinese tokens is very large and the TF-IDF scores can not be used to calculate the importance score directly.
Thus, we normalize the TF-IDF score into the range of 1 to 1000.

Besides, the multilingual NLP library HanLP~\cite{he-choi-2021hanlp} is used to classify Chinese tokens into their parts of speech and extract Chinese keywords from sentences.

\section{Details of G-Eval for Human Evaluation}
\label{appendix: geval}
G-Eval is a prompt-based evaluation method.
The first part of it is the prompt that contains the definition of the evaluation task and the evaluation criteria.
The second part is a chain-of-thoughts (CoT) which is a set of intermediate instructions generated by the LLM describing the detailed evaluation steps.
G-Eval concatenates the prompt, the evaluation steps of CoT, the dialogue history and the response as the instruction,
then call the LLM to output a score from 1 to 5 for each evaluation aspect, based on the defined criteria.

The concatenated form-filling paradigm is provided as below:

\textit{Task:}

\textit{Please make sure you read and understand the below instructs carefully. You will be provided a dialogue history and a response, you need to evaluate the \{informativeness/coherence/humanness\} between them and output a score of 1 to 5. The detailed evaluation criteria are provided below.
}

\textit{Evaluation Criteria: }

\textit{\{\} }

\textit{Evaluation Steps:}

\textit{1. Read the above task definition and the evaluation criteria carefully.
}

\textit{
2. Read the below given dialogue history and response carefully.
}

\textit{
3. Assign a score for \{informativeness/coherence/humanness\} on a scale of 1 to 5, where 1 is the lowest and 5 is the highest based on the Evaluation Criteria.
}

\textit{Dialogue history:}

\textit{\{\} }

\textit{Response:}

\textit{\{\} }

\textit{Score:}

The evaluation criteria of informativeness, coherence and humanness are as below:

\begin{itemize}
    \item Informativeness: Informativeness is used for evaluating whether the response contains the semantic relevant information in the dialogue history. 1 score means the response just repeats the dialogue history and fails to provide additional information, and 5 score means the response has appropriate and correct information.
    \item Coherence: Coherence is used for evaluating whether the response is relevant and consistent with the dialogue history. 1 score means the response is not suitable or is inconsistent with the dialogue history, 5 score means the response is suitable and consistent with the context.
    \item Humanness: Humanness is used for evaluating whether the response is similar to the tones of human beings. 1 score means the response is unnatural and the speaker seems not human-like, and 5 score means the response is natural and the speaker seems like a human being.
\end{itemize}



\bibliographystyle{cas-model2-names}

\bibliography{cas-refs}

\end{document}